\documentclass[journal]{vgtc}

\renewcommand{\manuscriptnotetxt}{}

\graphicspath{{Pictures/}{./}}

\usepackage{booktabs}
\usepackage{multirow}
\usepackage{array}
\usepackage{amsmath}
\usepackage{amssymb}
\usepackage{mathptmx} 
\usepackage{balance}  


\newcommand{\method}{Mover360}

\title{\method{}: Controllable Object Manipulation in 360$^\circ$ Panoramic Images}

\author{Haoyi Zhong, Fang-Lue Zhang*, Andrew Chalmers, and Taehyun Rhee}

\authorfooter{
  \item Haoyi Zhong and Andrew Chalmers are with Victoria
        University of Wellington, New Zealand.
        E-mail: \{haoyi.zhong, andrew.chalmers\}@vuw.ac.nz.
   \item Fang-Lue Zhang is with University of New South Wales, Australia. Email: fanglue.zhang@unsw.edu.au.
  \item Taehyun Rhee is with The University of Melbourne, Australia.
        E-mail: taehyun.rhee@unimelb.edu.au.
  \item *Fang-Lue Zhang is the corresponding author.
}

\abstract{%
We present \method{}, a controllable object manipulation framework for 360$^\circ$ images. Unlike perspective images, 360$^\circ$ images in equirectangular projection (ERP) exhibit horizontal wrap-around, latitude-dependent distortion, and global scene continuity, which makes object-level edits difficult for existing perspective editors to produce and for users to specify. To address this, \method{} centers on object \emph{Translation} (relocating a specified object within an existing panorama) while supporting reference-guided \emph{Insert} and \emph{Remove} as auxiliary tasks. Its interface unifies point-, bbox-, and mask-guided control by encoding each task into a fixed prompt and a compact, ERP-aligned instruction map. In the default point mode, a single click relocates an object, allowing the model to infer a plausible size, support, and illumination using panoramic context and an auxiliary depth condition. Structurally, \method{} is a lightweight adaptation of a pretrained diffusion transformer.
To generate paired supervision, we construct a UE5 data-generation pipeline with surface-aware object placement and randomized illumination, yielding large-scale paired data and a dual-domain benchmark of synthetic and real panoramas with ground truth for all three tasks. Across both test domains and two evaluation protocols, \method{} outperforms strong baselines for perspective editing, insertion, and inpainting in reconstruction fidelity, semantic consistency, and distributional quality.
Code and our benchmark dataset are available at \url{https://zhonghaoyi.github.io/Mover360/}.
}


\teaser{
  \centering
  \includegraphics[width=0.95\linewidth]{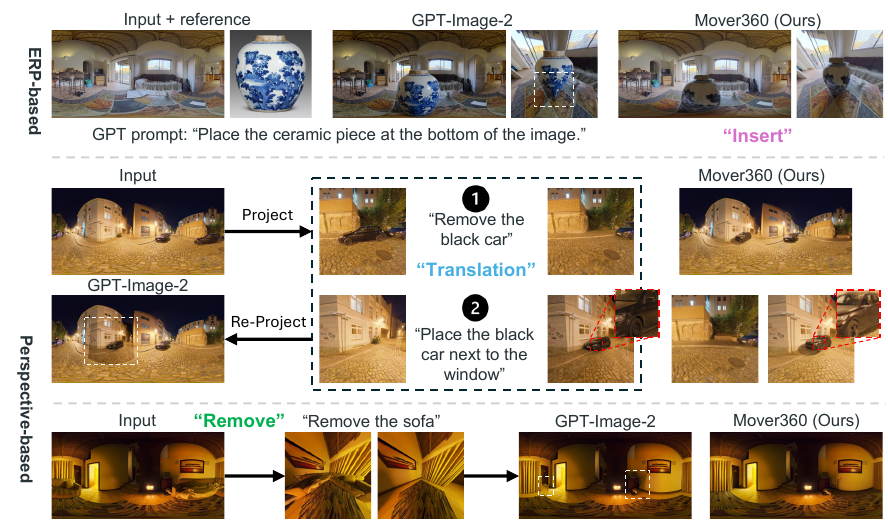}
  \caption{\textbf{Qualitative comparisons against a state-of-the-art general-purpose editor under two regimes.}  \emph{Top (ERP-based):} Given the full panorama, GPT-Image-2~\cite{openai2026gptimage2} inserts objects at an implausible scale, ignoring equirectangular geometry. \emph{Bottom (perspective-based):} Projecting, editing, and re-projecting a local perspective view discards panoramic context. For Translation, the re-projected boundary leaves a visible appearance discontinuity, and the relocated car lacks the specular highlights and cast shadows generated by \method{}. For Remove, content outside the perspective view survives re-projection, and unrelated regions drift (dashed boxes). \method{} produces globally consistent edits across all cases.}

  \label{fig:teaser}
}

\begin{document}

\firstsection{Introduction}

\maketitle


\providecommand{\method}{Mover360}
\setlength{\emergencystretch}{2em}

\label{sec:introduction}%
360$^\circ$ panoramas are widely used in virtual reality \cite{wang2020vr,wang2026rl}, immersive telepresence \cite{zhang2023survey}, indoor scene capture\cite{chen2022casual}, virtual staging \cite{shah2025virtual}, and environment lighting \cite{zhao2021adaptive,Zhao2024CVM}. In these applications, users often need object-level manipulation rather than global image generation: moving a piece of furniture to a new location, removing an unwanted object, or inserting a reference object into an existing scene. A practical panorama editor should preserve the surrounding environment while producing plausible object placement, scale, support, occlusion, and illumination.

Object manipulation in equirectangular projection (ERP) panoramas is substantially different from editing perspective images. ERP images represent a spherical scene on a rectangular grid: the horizontal axis is periodic, the left and right image boundaries correspond to adjacent longitudes, and the vertical axis introduces latitude-dependent distortion. As a result, an edit that appears local in image space may still affect the continuity of the global spherical scene. Directly applying perspective-image editing models to ERP panoramas can lead to seam artifacts, distorted object geometry, unstable scale, and inaccurate spatial control, especially when the edited object moves across a large region or lies near the panorama boundary.

A seemingly natural workaround is to project the edit region onto a perspective view, apply a perspective editing model there, and re-project the edited view back into the panorama. However, this projection pipeline has structural limits of its own, independent of how strong the underlying editor is. First, a pinhole view covers only a bounded field of view. Effects of the edit that extend beyond the view frustum, most typically the long cast shadow of a removed or relocated object, are neither observed nor updated, and re-projecting the edited frustum into the panorama tends to leave residual appearance discontinuities along the frustum boundary. Second, the editor perceives only the viewport rather than the surrounding sphere, so view-dependent appearance driven by out-of-view content cannot be synthesized; conversely, the influence of the edited object on the environment outside the viewport (its cast shadow, its reflections, and other illumination changes) is never generated. Figure~\ref{fig:teaser} illustrates both regimes with GPT-Image-2~\cite{openai2026gptimage2}, a state-of-the-art general-purpose editor: applied directly to the ERP panorama it ignores the equirectangular scene geometry, and applied through the perspective projection pipeline it exhibits exactly these frustum and context defects.

Recent image editing methods have made significant progress in instruction-guided editing, spatially conditioned generation, drag-based manipulation, reference-guided insertion, and object removal \cite{brooks2023instructpix2pix,zhang2023controlnet,shi2024dragdiffusion,chen2024anydoor,yang2023paintbyexample,suvorov2022lama,yu2025omnipaint}. Meanwhile, panorama generation and translation methods have studied how to adapt generative models to wide field-of-view images and ERP-specific artifacts \cite{wu2023panodiffusion,wang2024stitchdiffusion,zhang2024panfusion,wang2025panT,zhong2025se360}. These methods provide important foundations, but controllable object-level manipulation in 360$^\circ$ panoramas remains underexplored. Existing editing and insertion models are usually designed for perspective images and do not explicitly model the periodic boundary, latitude-dependent distortions, or the need for panorama-aligned spatial controls.

We present \method{}, a unified framework for controllable object manipulation in 360$^\circ$ panoramas. The framework centers on \emph{Translation}—the task of relocating a specified object within a scene. This is the most demanding of the practical panorama edits, as a single operation must simultaneously complete the background at the source, place the object plausibly at the target, and adjust its scale, occlusion, contact, and illumination to be globally consistent. The two components of this task, \emph{Remove} and \emph{Insert}, are also exposed as standalone functions through the same model and interface.
Each task is specified by a fixed prompt and a compact three-channel instruction map aligned to the equirectangular projection (ERP), encoding a source region, a target region, and a Gaussian target point. In the default point mode, a single click suffices: the user indicates only the desired contact location, and the model infers a plausible size, support, and illumination from panoramic context; bbox- and mask-guided variants offer explicit control when desired (Section~\ref{sec:method}).

The model builds on Flux2-Klein-4B-Base \cite{blackforestlabs2026flux2klein}, a pretrained rectified-flow transformer adapted with LoRA \cite{hu2022lora}: the instruction map, an auxiliary DA$^2$ depth map \cite{li2025da2}, and an optional reference image are encoded as ERP-aligned condition latents and routed through the backbone via its native token-coordinate convention \cite{su2024roformer}, with horizontal circular padding preserving left-right continuity \cite{zhang2024panfusion,zhong2025se360}. The model-side contribution is thus an adaptation scheme rather than a new backbone: it turns a perspective image generator into a unified panorama object editor without any architecture modification (Section~\ref{sec:method}).

A key challenge is the lack of paired supervision for object-level panorama editing, since real scenes are rarely observed both before and after a controlled edit. We address this with a UE5 data-generation pipeline and a dual-domain test benchmark of synthetic and real captured tuples, detailed in Section~\ref{sec:dataset}.

Our contributions are summarized as follows:
\begin{itemize}
    \item \textbf{Paired data construction for panorama object manipulation.} A UE5 generation pipeline with surface-aware placement, size-dependent motion paths, randomized illumination, and multi-view reference capture, yielding 12{,}150 paired camera--object sequences (60{,}750 training pairs per epoch) and a dual-domain benchmark of 210 synthetic and 50 real captured tuples with ground truth for all three tasks.
    \item \textbf{A translation-centered editing interaction for ERP panoramas.} A single unified model in which Translation is primary and Remove/Insert are auxiliary, driven by a three-channel ERP-aligned instruction map whose guidance scales from a single click to bbox and mask control, and realized by a minimally invasive multi-condition adaptation of a pretrained rectified-flow transformer.
    \item \textbf{Protocols and evidence.} Two complementary protocols for evaluating perspective editors on panoramas (direct ERP application and a strictly favorable perspective re-projection), under which \method{} achieves the strongest overall results on both test domains, with ablations isolating the depth condition and the instruction granularity.
\end{itemize}

\section{Related Work}
\label{sec:related_work}

\subsection{Image Editing and Object Manipulation}

Instruction-guided and controllable image editing has progressed from text-only commands to models that combine language, visual exemplars, masks, points, and structural controls. Instruction-tuned and recent multimodal editors improve command following and background preservation \cite{brooks2023instructpix2pix,geng2024instructdiffusion,huang2024smartedit,xu2025insightedit}, while ControlNet and drag-based methods provide stronger spatial guidance for local manipulation \cite{zhang2023controlnet,shi2024dragdiffusion,wu2024draganything}. Object-level editing further requires identity preservation together with plausible support, occlusion, shadow, reflection, and background completion. Reference-guided insertion, object-oriented inpainting, and recent object movement methods address these issues in perspective images or videos \cite{yang2023paintbyexample,chen2024anydoor,song2025insertanything,yu2025omnipaint,yu2025objectmover}. Although existing approaches to estimating environment maps from 2D images can potentially provide omnidirectional lighting information \cite{zhao2024salenet,zhao2025anisotropic}, current 2D editing methods are not designed to explicitly utilize environment maps to illuminate objects. ObjectMover is especially related to our Translation task because it studies realistic object displacement with synthetic game-engine supervision. However, these methods do not explicitly handle ERP periodicity, latitude-dependent distortion, or spherical scene continuity. \method{} instead performs Translation, Remove, and Insert directly on ERP panoramas with panorama-aligned source, target, point, and reference controls.

\subsection{360$^\circ$ Panorama Generation and Editing}

Early 360$^\circ$ content editing methods decompose ERP images into intrinsic content layers \cite{xu2024intrinsic,kou2024neural}, enabling straightforward modifications to lighting and object appearance. Recently, diffusion models have been widely leveraged in panorama generation to produce wide field-of-view imagery through ERP-aware, multi-view, or spherical-processing strategies\cite{wu2023panodiffusion,wang2024stitchdiffusion,zhang2024panfusion,ye2024diffpano,sun2025smgd}. Researchers have also explored learning inherent 3D representations from panoramic content to synthesize novel views \cite{kou2025omniplane,kou2026omniprior}. More recent work moves from panorama generation toward panorama translation and editing. 360PanT explores training-free text-driven panorama-to-panorama translation \cite{wang2025panT}; Omni$^2$ unifies omnidirectional generation and editing with Any2Omni \cite{yang2025omni2}; SE360 constructs hierarchical editing pairs for multi-condition object editing in panoramas \cite{zhong2025se360}; and World-Shaper studies ERP-domain geometry-aware panoramic editing with generate-then-edit supervision \cite{liang2026worldshaper}. These works show the importance of boundary continuity and spherical geometry, but they mainly emphasize generation, semantic editing, or generation-editing unification. Our focus is narrower and more controllable: moving, removing, or inserting a specific object in an existing panorama with explicit object-level spatial guidance. As detailed in Section~\ref{sec:baselines}, these systems cannot serve as baselines for this setting: they either expose no object-level spatial control, edit only through global text instructions, or do not release the relevant weights.

\subsection{Editing Data Construction}

Large-scale paired supervision is difficult to collect for real object manipulation because the same scene must be observed before and after a controlled edit. Existing editors therefore rely on synthetic instructions, generated edit pairs, game-engine rendering, or automatically constructed panorama editing data \cite{brooks2023instructpix2pix,xu2025insightedit,yu2025objectmover,yang2025omni2,zhong2025se360,liang2026worldshaper}. Our data pipeline follows this need for controlled supervision but targets a different granularity: it renders surface-aware object trajectories in UE5, provides masks and background frames, randomizes global and local illumination, and captures multi-view references so that Translation, Remove, and Insert can be trained within one panorama object-manipulation framework.

\section{\method{} Dataset and Benchmark}
\label{sec:dataset}

\subsection{Paired Panorama Editing}
\begin{figure*}[t]
\centering
\includegraphics[width=0.96\textwidth]{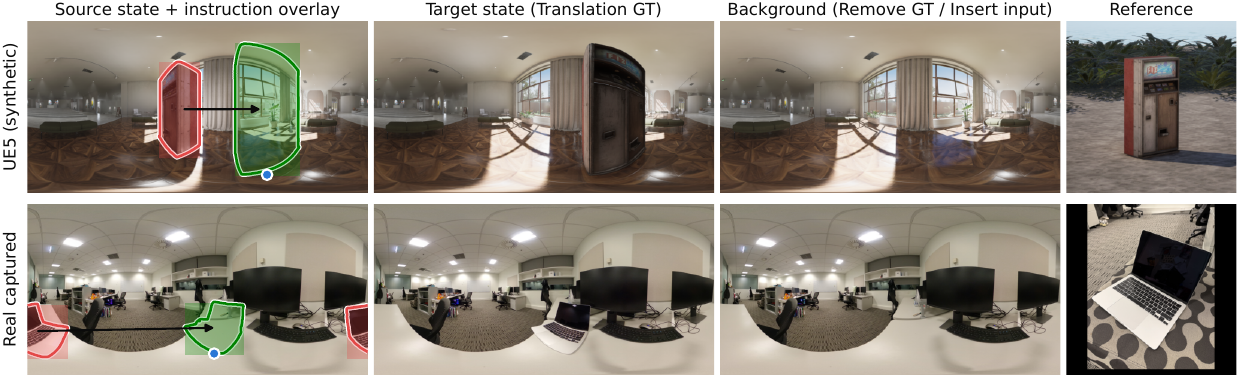}
\caption{Anatomy of one \method{} editing tuple from the UE5 test set (top) and from the real captured test set (bottom). Each tuple contains a source-state panorama, a target-state panorama, an object-absent background panorama, and a perspective reference image, and therefore supervises all three tasks: Translation (source$\,\to\,$target), Remove (source$\,\to\,$background), and Insert (background$\,+\,$reference$\,\to\,$object present). The left column overlays all three guidance granularities of the instruction map: fine object masks (contours), bbox regions (translucent rectangles), and the Gaussian target point placed at the bottom-center of the target box (blue dot), with the source in red and the target in green. Note that the source object of the real example straddles the horizontal wrap-around boundary of the ERP.}
\label{fig:dataset_samples}
\end{figure*}

Training \method{} requires paired panoramas in which the same scene is observed before and after a controlled object-level change. We construct such data in UE5 using five scenes, 608 unique objects (926 scene-level object instances), 60 predefined motion paths, three panoramic cameras per path, and 10-frame trajectories per camera; five frames per trajectory are retained by stratified sampling to reduce temporal redundancy, and degenerate frames are filtered out. Each retained frame provides a $2048 \times 1024$ RGB ERP panorama and an aligned object mask, used at $1024 \times 512$ for training. The resulting corpus contains 12{,}150 valid camera--object sequences. The supported tasks are derived from the same rendered trajectories, as illustrated in Figure~\ref{fig:dataset_samples}: frame pairs supervise Translation, object-to-background pairs supervise Remove, and background-to-object pairs with a reference image supervise Insert. We dynamically sample valid pairs during training, producing 60{,}750 samples per epoch (36{,}450 Translation, 12{,}150 Remove, and 12{,}150 Insert pairs).

\subsection{Surface-Aware Object Motion}
The 60 paths are object-size-dependent rather than uniformly scaled from one template. We divide them into small, medium, large, and extra-large categories, and each object is assigned only to paths compatible with its size and available motion space. This avoids unrealistic samples such as a large object moving through a narrow path or a small object being displaced across an excessively large region. Objects are placed by surface raycasting to avoid penetration and maintain contact with floors, walls, ceilings, or slanted planes. On slanted surfaces, the object's up direction is aligned with the detected surface normal, producing physically plausible support throughout the trajectory.

Each path is observed by three panoramic cameras. The cameras share the same object motion but view it from different positions and orientations, so the same translation appears with different poses, locations, and ERP distortions. For each camera, we render a 10-frame object sequence and a one-frame background sequence. The object sequence provides translation pairs and object-present targets, while the background frame provides object-absent supervision for Remove and object-absent inputs for Insert.

\begin{figure}[t]
\centering
\includegraphics[width=\columnwidth]{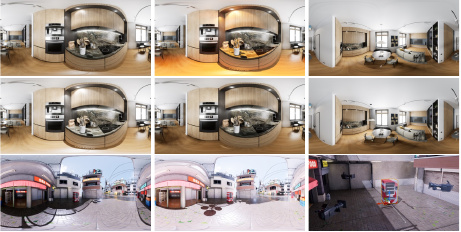}
\caption{Examples from the \method{} data generation pipeline. The top two rows show randomized nearby lamps, including variations in intensity, temperature, scattering or beam angle, and position. In the third row, the first two examples show outdoor/global light-source variation, and the last example shows one object path captured by three panoramic cameras.}
\label{fig:light_change}
\end{figure}

\subsection{Lighting and Reference}
We randomize illumination at both global and local levels. For each path rendering instance, including three camera sequences for a size-compatible object, we resample the main light and SkyLight intensity and angles. Auxiliary RectLights are scaled according to their initial ratios so that they remain consistent with the global light change. We also randomize nearby lamps, including position, intensity, temperature, and beam or scattering angle, as illustrated in Figure~\ref{fig:light_change}.

For reference-guided insertion, each object is additionally rendered from eight perspective viewpoints under diverse illumination. To avoid unstable training from extreme viewpoint mismatch, we compute DINOv3 feature similarity \cite{simeoni2025dinov3} between each candidate reference view and the target object crop in the panorama, retain the top three views, and randomly sample one of them. For translation, we sample two frames from the same camera sequence with a temporal interval of at least two frames. Each rendered camera-object sequence contributes three translation pairs per epoch, while Remove and Insert each contribute one sampled pair.

\subsection{Test Benchmark and Statistics}
\label{sec:dataset_stats}

\noindent\textbf{Dual-domain test benchmark.}
For evaluation, we complement the training corpus with a two-part benchmark; Figure~\ref{fig:dataset_samples} shows one editing tuple from each part. The UE5 test set contains 210 tuples covering 48 objects, rendered with the same pipeline as the training corpus but from held-out sequences: no test camera--object sequence appears in the training index. The real test set contains 50 tuples covering 15 unique objects captured in real scenes: for each tuple, we photograph the panorama before and after physically relocating an object, an object-absent background panorama, and a perspective reference photo of the object. Every tuple therefore provides ground truth for all three tasks in both domains, enabling paired quantitative evaluation on real data rather than reference-free scoring.

\begin{figure*}[t]
\centering
\includegraphics[width=0.92\textwidth]{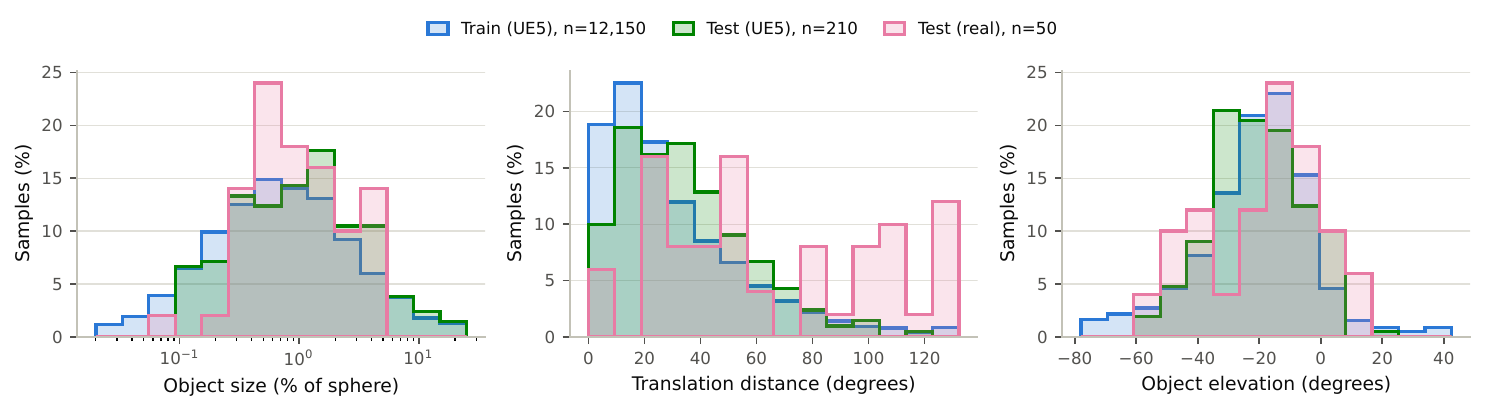}
\caption{Distributions of sphere-aware statistics for the \method{} training corpus and the two test sets: object size as a percentage of the viewing sphere (log scale), great-circle translation distance, and object elevation. The two synthetic sets share similar distributions with a wider size range, while the real test set deliberately stresses long-range translations (median $50.9^\circ$, with a quarter of the moves beyond $103^\circ$).}
\label{fig:dataset_stats}
\end{figure*}

\noindent\textbf{Sphere-aware statistics.}
Table~\ref{tab:dataset_stats} and Figure~\ref{fig:dataset_stats} summarize the training corpus and both test sets using statistics computed from the ERP object masks on the sphere: object size as the solid-angle fraction of the viewing sphere, translation distance as the wrap-aware great-circle displacement of the mask centroid, and object elevation as the centroid latitude. Three properties stand out. First, object sizes span more than two orders of magnitude (roughly $0.1$--$27\%$ of the sphere), with a heavy tail of large near-field objects; the largest objects cannot be framed by any single pinhole crop, which motivates the field-of-view fallback of the perspective evaluation protocol in Section~\ref{sec:baselines}. Second, translations are genuinely panoramic: the median great-circle displacement is $23.4^\circ$ in training and $31.4^\circ$/$50.9^\circ$ on the synthetic/real test sets, and a quarter of the real moves exceed $103^\circ$. Such displacements leave any fixed perspective crop and substantially change the local ERP distortion between the source and the target, which is precisely the regime that motivates panorama-native translation. The real benchmark is deliberately harder than the training distribution in this respect, probing generalization to long-range rearrangements rather than in-distribution interpolation. Third, elevations concentrate below the horizon, reflecting objects resting on floors and support surfaces, while the training corpus additionally covers wall- and ceiling-mounted objects at elevations up to $+42^\circ$.

\begin{table}[t]
\centering
\scriptsize
\caption{Statistics of the \method{} training corpus and the dual-domain test benchmark. Distribution rows report median [interquartile range], computed from the ERP object masks: object size is the solid-angle fraction of the sphere, translation is the wrap-aware great-circle displacement of the mask centroid (sampled training pairs for the corpus, source-to-target pairs for the test sets), and elevation is the centroid latitude.}
\label{tab:dataset_stats}
\resizebox{\columnwidth}{!}{%
\begin{tabular}{lccc}
\toprule
 & Train (UE5) & Test (UE5) & Test (real) \\
\midrule
Editing samples & 12{,}150 seq. & 210 & 50 \\
Sampled pairs per epoch & 60{,}750 & -- & -- \\
Unique objects & 608 & 48 & 15 \\
Ground truth for & \multicolumn{3}{c}{Translation / Remove / Insert} \\
\midrule
Object size (\% of sphere) & 0.69 [0.27, 1.72] & 1.04 [0.40, 2.15] & 0.79 [0.49, 1.72] \\
Translation ($^\circ$) & 23.4 [11.9, 42.3] & 31.4 [18.0, 47.5] & 50.9 [34.2, 103.0] \\
Elevation ($^\circ$) & $-18.9$ [$-30.5$, $-9.7$] & $-20.6$ [$-30.7$, $-10.3$] & $-14.5$ [$-37.7$, $-7.0$] \\
\bottomrule
\end{tabular}}
\end{table}

\section{Method}
\label{sec:method}

\begin{figure*}[t]
\centering
\includegraphics[width=0.88\textwidth]{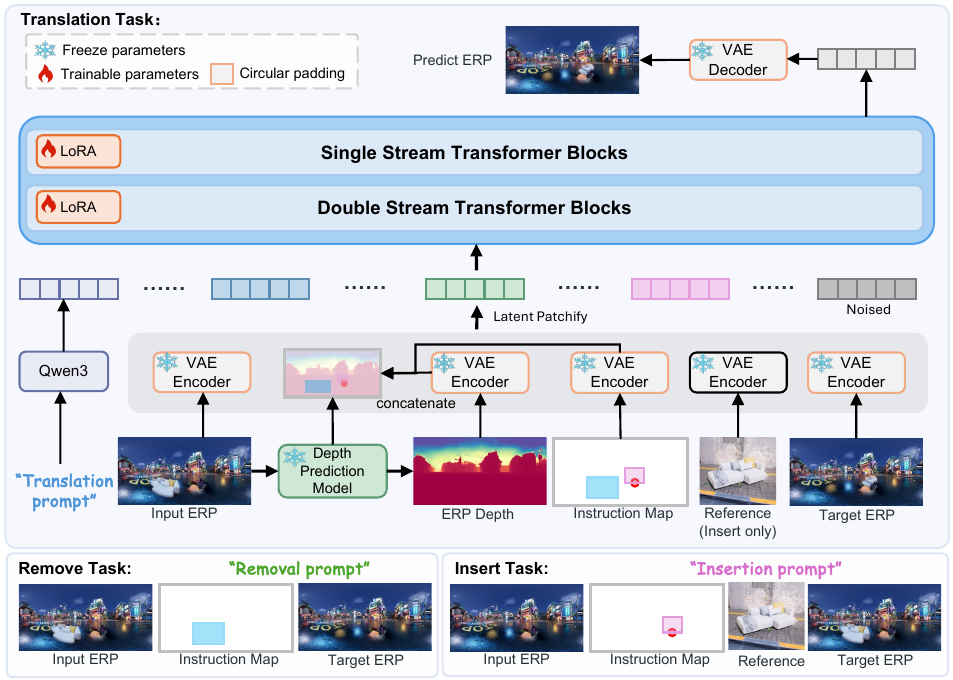}
\caption{Overview of \method{}. The framework takes an input ERP panorama, a fixed task prompt, an ERP-aligned instruction map, an auxiliary DA$^2$ depth map, and an optional reference image for insertion. Condition latents are concatenated and patchified before being processed by the Flux2-Klein diffusion transformer. We adapt the native 4D RoPE coordinates to distinguish text tokens, noisy target panorama tokens, and multiple conditioning image slots. LoRA adapters are trained while the reference branch is disabled for translation and remove. VAE encoding and decoding are wrapped with horizontal circular pad-and-unpad to better respect panorama periodicity.}
\label{fig:pipeline}
\end{figure*}

\subsection{Overview}
\label{sec:method_overview}

Figure~\ref{fig:pipeline} shows the architecture of \method{}. The framework is built on Flux2-Klein-4B-Base \cite{blackforestlabs2026flux2klein} and operates in the latent space of its VAE. Given an input ERP panorama, we first estimate an auxiliary depth map using a frozen DA$^2$ model \cite{li2025da2}. The input panorama, instruction map, normalized ERP depth, optional reference image, and training target are encoded by the VAE encoder. The condition latents are concatenated and patchified into token sequences before being passed to the diffusion transformer. We further adapt the native Flux2 4D rotary position encoding so that the transformer can distinguish text tokens, noisy target panorama tokens, and multiple conditioning image slots while preserving the 2D spatial layout of each image-like input. The transformer predicts the flow-matching velocity \cite{lipman2023flowmatching} of the target latent, and the VAE decoder reconstructs the predicted ERP panorama.

\method{} uses the same backbone for all three tasks. The difference lies in the active condition branches and instruction-map channels. Translation uses the input panorama, the source/target control map, the fixed translation prompt, and auxiliary depth. Remove uses the input panorama, the source-region control map, the fixed Remove prompt, and auxiliary depth. Insert additionally activates the reference branch, which encodes a single reference object image; for Translation and Remove, the reference branch is directly disabled.

\subsection{Task Formulation}
\label{sec:task_formulation}

Given an input ERP panorama $I \in \mathbb{R}^{H \times W \times 3}$, \method{} predicts an edited panorama $\hat{I}$ according to a task-specific condition:
\begin{equation}
    \hat{I} = f_\theta(I, P, C, D, R),
\end{equation}
where $P$ is a fixed task prompt, $C$ is an ERP-aligned instruction map, $D$ is an auxiliary ERP depth map estimated by DA$^2$, and $R$ is an optional reference image used only for insertion. We consider three tasks: Translation, Remove, and Insert. Translation is the primary task: it composes the other two (background completion at the source region and plausible object placement at the target) within a single consistent edit, and it is where panorama-specific reasoning about projected size, support, and distortion is most demanding. Remove and Insert are trained jointly as auxiliary tasks; they share the backbone and the instruction interface, regularize the two halves of Translation, and remain useful edits in their own right.

The instruction map $C \in \mathbb{R}^{H \times W \times 3}$ contains three channels: $  C = [C_s, C_t, C_p]$
where $C_s$ encodes the source region, $C_t$ encodes the target region, and $C_p$ encodes the target point. The source and target regions can be represented either as full object masks or as rectangular box masks. The target point is represented by an isotropic Gaussian heatmap centered at the desired contact location, with a standard deviation of 5 pixels at the training resolution. A point specifies where the object should contact or be placed, while a bbox or mask can additionally specify the desired spatial extent. Unused channels are set to zero according to the task.

For Translation, $C_s$ marks the source object and $C_t$ or $C_p$ specifies the target location. For Remove, only $C_s$ is active and marks the object to remove. For Insert, the target region or target point is provided through $C_t$ or $C_p$, and the reference image $R$ specifies the appearance of the object to insert.

Each task uses one fixed task prompt. The Translation prompt is: \emph{``Image 1 is the panorama to edit. Image 2 marks the source and target regions. Move the object from the source region to the target region.''} The Remove prompt is: \emph{``Image 1 is the panorama to edit. Image 2 marks the object to remove. Remove the object from the source region.''} The Insert prompt is: \emph{``Image 1 is the panorama to edit. Image 2 is the object reference. Image 3 marks the target region. Add the object at the target region.''}

\subsection{ERP-Aligned Instruction Map}
\label{sec:instruction_map}

The instruction map provides the primary spatial control for object manipulation. Its three channels are aligned with the ERP image grid and are encoded at the same panorama resolution before VAE encoding. The source-region channel is used for Translation and Remove to localize the object being manipulated. The target-region channel is used when a dense or box-shaped target region is available. The target-point channel is a lightweight alternative that marks the desired target location with a Gaussian point.

The default interface is point-guided. This design reduces user burden in panoramic editing: a non-expert user often does not know how large the target object should appear after translation or insertion, because its projected size changes with scene depth, vertical latitude, and the local contact surface. Instead of asking the user to draw the exact target mask, point guidance asks only for a contact or placement point. The model then predicts the object extent, scale, and support relationship from the input panorama, the instruction map, and the auxiliary depth condition.

We also support bbox-guided control by representing a rectangular target region in $C_t$. This stronger control is useful when users want to customize the generated object's size or occupy a specified target area. Since bbox guidance provides a more explicit spatial prior than a single Gaussian point, it can improve quantitative reconstruction metrics. We therefore use point guidance as the default lightweight interface and report bbox guidance as a stronger-control variant in the ablation study.

\subsection{Auxiliary ERP Depth Conditioning}
\label{sec:depth_condition}

We use DA$^2$ \cite{li2025da2} to estimate an ERP depth map from the input panorama. The same predicted-depth pipeline is used during both training and inference. We do not use rendered depth targets or an auxiliary depth prediction objective. The depth map is used only as an additional geometric condition that helps the editor reason about object support, scene layout, and scale. During training, this depth condition is stochastically dropped with probability $0.1$, which encourages the model to remain robust when the auxiliary depth cue is unreliable or unavailable.

The predicted depth is normalized before VAE encoding. Let $D$ be the predicted depth map. We compute quantile statistics only over valid depths,
\begin{equation}
    \Omega_D = \{(i,j) \mid \operatorname{isfinite}(D_{ij}) \wedge D_{ij} > 0\}.
\end{equation}
The lower and upper normalization bounds are
\begin{equation}
    d_{\mathrm{lo}} = Q_{0.00}(D_{\Omega_D}), \qquad
    d_{\mathrm{hi}} = Q_{0.98}(D_{\Omega_D}),
\end{equation}
and the normalized depth is
\begin{equation}
    D_{\mathrm{norm}} =
    \operatorname{clip}\left(
    \frac{D-d_{\mathrm{lo}}}{d_{\mathrm{hi}}-d_{\mathrm{lo}}},
    0, 1
    \right).
\end{equation}
Finally, the VAE input is mapped to $[-1,1]$ and repeated across three channels:
\begin{equation}
    D_{\mathrm{vae}} = \operatorname{repeat}_3(2D_{\mathrm{norm}} - 1).
\end{equation}
Using predicted depth for both synthetic and real inputs keeps the training and inference pipelines consistent, since rendered UE5 depth is not available for real captured panoramas at test time.

\subsection{Reference Branch for Insertion}
\label{sec:reference_branch}

The reference branch is activated only for the Insert task. The input reference image is a single perspective image that specifies the appearance of the object to be inserted. It is encoded by the VAE encoder and fused with the other condition latents. For Translation and Remove, this branch is directly disabled rather than filled with a null reference image.

During training, each object has eight candidate perspective reference views rendered under diverse illumination. A randomly selected view can be unstable when it differs too much from the object appearance in the target panorama. We therefore compute DINOv3 feature similarity \cite{simeoni2025dinov3} between each candidate reference view and the target object crop in the panorama, select the top three most similar views, and randomly sample one of them for training. This strategy maintains reference-view diversity while avoiding extreme reference-target viewpoint mismatch.

\subsection{Latent Encoding and Patchification}
\label{sec:latent_encoding}

All image-like inputs are encoded in the latent space of Flux2-Klein-4B-Base. Let $E_{\mathrm{vae}}$ denote the VAE encoder. The conditioning latents are computed as
\begin{equation}
\begin{aligned}
    z_I &= E_{\mathrm{vae}}(I),
    & z_C &= E_{\mathrm{vae}}(C), \\
    z_D &= E_{\mathrm{vae}}(D_{\mathrm{vae}}),
    & z_R &= E_{\mathrm{vae}}(R),
\end{aligned}
\end{equation}
where $z_R$ is used only for insertion. During training, the ground-truth target ERP $I^\star$ is also encoded into a target latent $z_0 = E_{\mathrm{vae}}(I^\star)$ and corrupted according to the flow-matching schedule. The condition latents are concatenated and patchified into token sequences before being passed into the diffusion transformer.

The fixed task prompt $P$ is encoded by the pretrained Qwen3 text branch of the backbone \cite{yang2025qwen3}. The resulting text tokens are used together with the patchified condition latents to guide the transformer prediction.

\subsection{Token-Coordinate Assignment for Multi-Condition Inputs}
\label{sec:rope_adaptation}

Flux2-Klein uses rotary position embeddings (RoPE) with 4D position IDs \cite{su2024roformer}. We follow this native design and represent each transformer token by a coordinate tuple $(T,H,W,L)$. We do not modify the RoPE formulation or the transformer architecture; we only assign coordinates within the native format so that the pretrained backbone can parse our multi-condition input, using different axes for text tokens, noisy panorama latent tokens, and conditioning image tokens.

For text tokens, only the sequence axis is used. For the $\ell$-th text token, the position ID is
\begin{equation}
    \mathrm{pos}_{\mathrm{text}}(\ell) = (0,0,0,\ell).
\end{equation}
For the noisy target panorama latent, we use the spatial latent grid. A noisy latent token at spatial position $(h,w)$ is assigned
\begin{equation}
    \mathrm{pos}_{\mathrm{noise}}(h,w) = (0,h,w,0).
\end{equation}
For conditioning image tokens, including the input panorama, optional reference image, instruction-map guidance, and depth guidance, we use the same spatial axes but assign each conditioning image slot a distinct $T$ coordinate:
\begin{equation}
    \mathrm{pos}_{\mathrm{cond}}(k,h,w) = (10k,h,w,0),
\end{equation}
where $k=1,2,3,\ldots$ denotes the conditioning image slot order. This separates the denoising latent tokens from different conditioning images while preserving the 2D spatial layout of each image-like condition.

The text RoPE embeddings and image/latent RoPE embeddings are computed separately through the Flux2 positional embedding module. They are then concatenated in the same order as the transformer token sequence:
\begin{equation}
    \mathcal{S}
    = \mathcal{S}_{\mathrm{text}}
    \Vert \mathcal{S}_{\mathrm{noise}}
    \Vert \mathcal{S}_{\mathrm{cond}} .
\end{equation}
Here $\mathcal{S}_{\mathrm{text}}$, $\mathcal{S}_{\mathrm{noise}}$, and $\mathcal{S}_{\mathrm{cond}}$ denote the text tokens, noisy panorama latent tokens, and conditioning image tokens, respectively. The concatenated RoPE embedding is injected into both the Flux2 double-stream transformer blocks and the single-stream transformer blocks. As a result, the transformer can jointly attend over text, the current noisy panorama latent, and multiple image conditions while still distinguishing their modality and slot identity through 4D coordinates.

For 360$^\circ$ panoramas, we do not introduce a new spherical RoPE. Horizontal continuity is handled at the VAE boundary by circularly padding the VAE inputs and cropping the corresponding outputs (Section~\ref{sec:circular_padding}), following panorama pad-and-unpad designs \cite{zhang2024panfusion,zhong2025se360}. RoPE is then applied on the resulting latent grid using the $H/W$ coordinates described above.

\subsection{Panorama-Aware VAE Circular Padding}
\label{sec:circular_padding}

ERP panoramas are periodic along the horizontal axis, since the left and right boundaries correspond to adjacent longitudes on the sphere. Standard zero padding in convolution layers breaks this periodicity and can introduce boundary artifacts. Circular padding is the classic remedy in panorama processing \cite{zhang2024panfusion,zhong2025se360}, and we adopt it unchanged rather than claim it as a contribution: the input panorama is circularly padded along the horizontal axis (64 pixels per side) before VAE encoding and the resulting latent is cropped back to the original width, and latents are symmetrically padded before VAE decoding with the decoded panorama cropped back. Latent and output sizes therefore remain unchanged. This wrapping is applied only around the VAE; the diffusion transformer operates on patchified latent tokens with the coordinate assignment described in Section~\ref{sec:rope_adaptation}.

\subsection{LoRA Adaptation}
\label{sec:lora}

We adapt Flux2-Klein-4B-Base with LoRA \cite{hu2022lora} rather than updating all model parameters. LoRA adapters are inserted into the attention query/key/value layers and projection layers of the transformer. We use rank 32, alpha 32, and dropout 0, resulting in 37,355,520 trainable LoRA parameters. This parameter-efficient adaptation preserves most pretrained weights while enabling the model to learn ERP-specific object manipulation behavior from paired data.

\subsection{Flow-Matching Training Objective}
\label{sec:training_objective}

\method{} follows the flow-matching objective \cite{lipman2023flowmatching} of the Flux2-Klein backbone. Let $z_0$ be the VAE latent of the target ERP panorama and let $\epsilon \sim \mathcal{N}(0,I)$ be Gaussian noise. Following our implementation, we sample $u \sim \mathcal{N}(0,1)$ and set $t=\sigma(u)$, where $\sigma(\cdot)$ is the sigmoid function. Thus, the timestep follows a sigmoid-normal, or logistic-normal, distribution rather than a uniform distribution. We then construct the interpolated latent
\begin{equation}
    z_t = (1-t) z_0 + t\epsilon .
\end{equation}
The target velocity is
\begin{equation}
    v = \epsilon - z_0 .
\end{equation}
Given the task prompt and condition latents $c$, the diffusion transformer predicts
\begin{equation}
    \hat{v} = v_\theta(z_t,t,c).
\end{equation}
The training objective is
\begin{equation}
    \mathcal{L}_{\mathrm{FM}} =
    \mathbb{E}_{z_0,\epsilon,u,c}
    \left[
        \|v_\theta(z_t,t,c)-v\|_2^2
    \right],
    \qquad t=\sigma(u).
\end{equation}
The optimization uses only this flow-matching objective; no additional RGB reconstruction, perceptual, seam, prompt, or auxiliary depth objective is used.

\section{Implementation Details}
\label{sec:implementation_details}

We train at an ERP resolution of $1024 \times 512$ using AdamW with an initial learning rate of $2 \times 10^{-4}$. The learning rate follows a step-wise CosineAnnealingLR schedule. Training uses 8 RTX Pro6000 GPUs with a per-GPU batch size of 2. The model is trained for 8 epochs, corresponding to approximately 30K iterations, and training takes about 20 hours. During training, the instruction map is sampled to cover all interfaces: an active region channel carries the fine object mask or its bounding box with equal probability, and the target guidance is exclusively a Gaussian point (70\%) or a target region (30\%). The task prompt is dropped with probability $0.01$ during training, which provides the empty-text prediction used as the negative branch of classifier-free guidance. At inference time, we generate results at $1024 \times 512$ resolution using 50 sampling steps and a text classifier-free guidance scale of 3, keeping the image conditions fixed in both guidance branches. A single panorama takes approximately 20 seconds to generate. All point-, bbox-, and mask-guided results in our experiments come from the same trained weights and differ only in the instruction supplied at inference; the w/o-depth ablation is a separately trained model.

\begin{figure*}[t]
\centering
\includegraphics[width=0.9\textwidth]{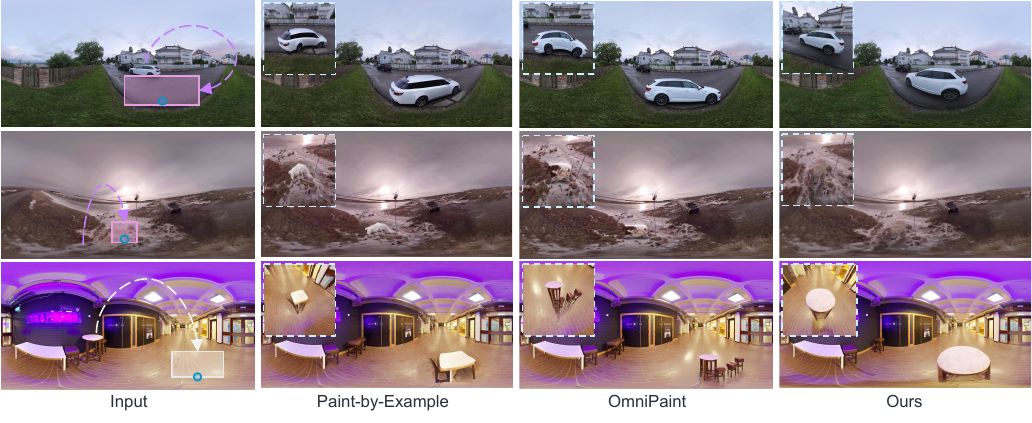}
\caption{Qualitative comparison for the Translation task. Baselines often leave residual content at the source location, distort the moved object, or produce less plausible contact with the target surface. \method{} uses the point or region instruction to move the object while preserving the surrounding panorama.}
\label{fig:move_compare}
\end{figure*}
\section{Experiments}
\label{sec:experiments}

\subsection{Evaluation Setup}

We evaluate \method{} on the dual-domain benchmark described in Section~\ref{sec:dataset_stats}, with 210 synthetic UE5 tuples and 50 real captured tuples per task, each providing paired ground truth. We report the two test sets separately, since they probe complementary aspects: the UE5 set provides exactly paired renders under controlled conditions, while the real set measures transfer to captured panoramas with real optics, lighting, and scene statistics and, as shown in Table~\ref{tab:dataset_stats}, stresses long-range translations beyond the training distribution.

\subsection{Baselines}
\label{sec:baselines}

\noindent\textbf{Why existing panorama methods are not directly comparable.}
Although a growing number of 360$^\circ$ generation and editing systems exist, none of them supports the controllable, object-level manipulation setting studied in this paper (mask-, bbox-, or point-guided editing of a specified object in an existing panorama) with publicly released weights. Panorama generation methods, including PanoDiffusion \cite{wu2023panodiffusion}, StitchDiffusion \cite{wang2024stitchdiffusion}, PanFusion \cite{zhang2024panfusion}, DiffPano \cite{ye2024diffpano}, and SMGD \cite{sun2025smgd}, synthesize panoramas from text or partial observations and provide no interface for manipulating an existing object. 360PanT \cite{wang2025panT} performs global panorama-to-panorama translation without localized object control, and Omni$^2$ \cite{yang2025omni2} edits only through global text instructions, without an object-level spatial-guidance interface and without released weights. SE360 \cite{zhong2025se360} is the closest system, performing multi-condition object editing on ERP panoramas; however, among our three tasks only its removal model is publicly released, while the weights of its reference-guided insertion model are not publicly available, and point- or bbox-guided translation of an existing object is not supported. World-Shaper \cite{liang2026worldshaper} is concurrent work without a public implementation at the time of writing. Consequently, no existing panorama editing system can be evaluated on the full Translation/Remove/Insert protocol with mask, bbox, or point guidance; the only partial exception is SE360's released removal model, which we include in the Remove comparison. All remaining baselines are publicly released perspective object editors.

We therefore compare with representative, publicly available perspective object-editing baselines according to task. For Translation, we evaluate DragAnything \cite{wu2024draganything}, AnyDoor \cite{chen2024anydoor}, Paint-by-Example \cite{yang2023paintbyexample}, OmniPaint \cite{yu2025omnipaint}, and Insert-Anything \cite{song2025insertanything}. For Insert, we compare with Paint-by-Example, AnyDoor, OmniPaint, and Insert-Anything. For Remove, we compare with LaMa \cite{suvorov2022lama}, OmniPaint, and the released removal model of SE360 \cite{zhong2025se360}. The baselines are applied using their corresponding input interfaces, while \method{} uses the same task prompts and instruction-map format described in Section~\ref{sec:method}.

\noindent\textbf{Baseline protocols.}
Since all compared editors are designed for perspective images, we adapt each baseline to its native interface; Table~S1 in the supplementary material summarizes the per-method protocols. For Translation, DragAnything is given a source handle at the object center and a target handle at the desired location with a linear drag trajectory, and the last generated video frame is taken as the moved image. Baselines without native object movement use a remove-then-insert protocol: OmniPaint first removes the source object, and the source object is then used as the reference for insertion at the target region. All baselines are provided with the required known-size target bbox or mask, while \method{} is evaluated by default with only a target point.

\noindent\textbf{ERP vs.\ perspective application.}
We apply each baseline under two complementary protocols. In the \emph{ERP protocol} (Table~\ref{tab:main_results}), the baseline operates directly on the equirectangular panorama (or on axis-aligned ERP crops for crop-based editors), which exposes it to ERP distortion it was never trained on. To factor this nuisance out, we additionally introduce a \emph{perspective protocol} (Table~\ref{tab:main_results_pers}): the panorama is projected to a square pinhole view centered on the edit region, with the field of view set adaptively to $\mathrm{clip}(1.2\times$ the region's angular extent$, 35^\circ, 150^\circ)$; the baseline edits natively inside this locally undistorted view, its target box is taken as the bounding rectangle of the \emph{projected} fine mask, and the entire edited view is re-projected into the panorama, leaving all pixels outside the view untouched. Re-projecting the full view rather than only the target box credits every effect a baseline produces beyond the box, such as the cast shadow of an inserted object or the shadows and reflections erased together with a removed one. Objects whose required field of view exceeds $150^\circ$ cannot be framed by a pinhole camera and fall back to the ERP protocol with the same compositing convention, and objects straddling the $\pm180^\circ$ seam are made contiguous by rolling the panorama before editing. Reference images for reference-guided baselines follow an object-on-white convention: the source object is projected to its own tightly framed perspective view and pixels outside its fine mask are set to white. This protocol removes ERP distortion from every baseline input and output path and is therefore strictly favorable to the baselines. \method{} operates natively on the ERP in both tables; for a fair comparison under the perspective protocol, its output is composited in the same way, with every pixel outside the perspective-view crop replaced by the input panorama, so only its full-panorama metrics (FAED and PSNR) differ from the ERP protocol.

\begin{figure*}[t]
\centering
\includegraphics[width=0.95\textwidth]{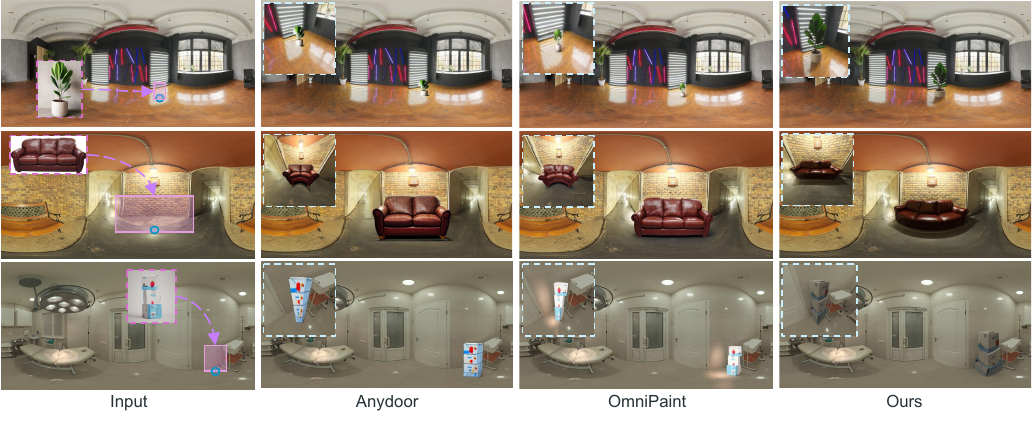}
\caption{Qualitative comparison for the Insert task. Reference-guided baselines may copy viewpoint-specific appearance or produce inconsistent object scale. \method{} better adapts the inserted object to the target region and the panorama context.}
\label{fig:insert_compare}
\end{figure*}

\begin{figure*}[t]
\centering
\includegraphics[width=0.95\textwidth]{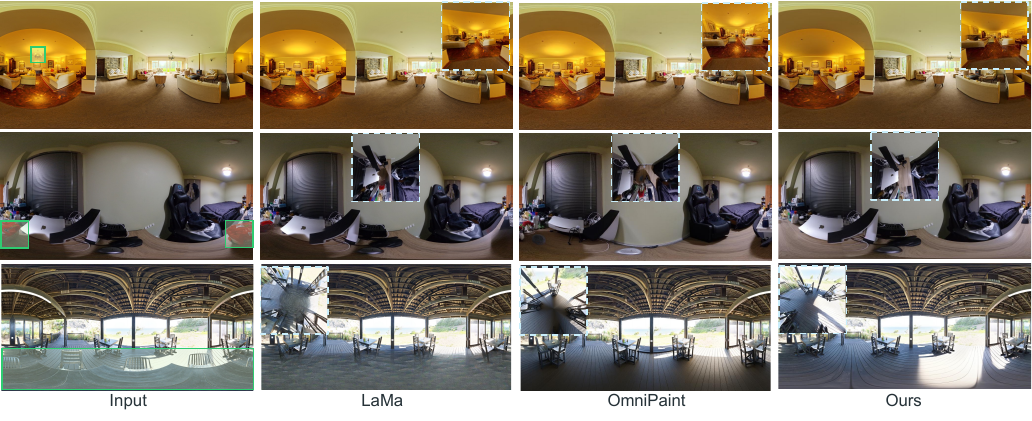}
\caption{Qualitative comparison for the Remove task. LaMa is a strong specialized inpainting baseline, but \method{} performs removal within the same unified framework used for translation and insertion.}
\label{fig:remove_compare}
\end{figure*}

\begin{figure*}[t]
\centering
\includegraphics[width=0.95\textwidth]{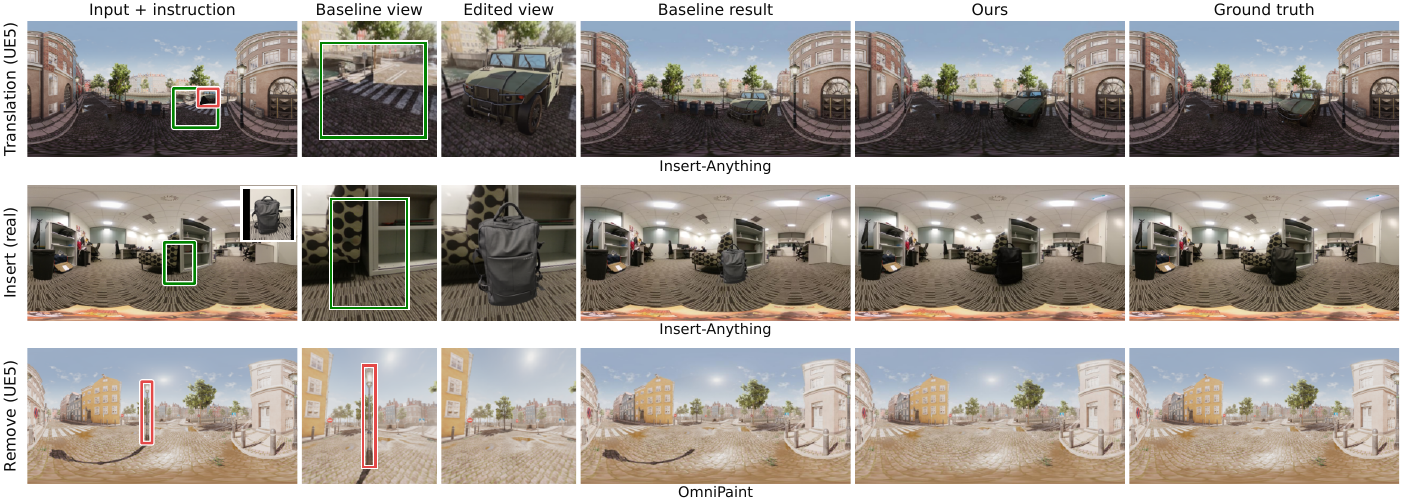}
\caption{Qualitative comparison under the \emph{perspective protocol}. Rows show Translation (UE5), Insert (real), and Remove (UE5); columns show the input panorama with the instruction regions overlaid (source in red, target in green, reference inset for Insert), the square perspective view given to the baseline with the provided target box, the baseline's edited view, the baseline's final panorama after re-projection, \method{} (mask), and the ground truth. For Translation, the baseline crops the source-removed panorama. Mask-conditioned baselines struggle with global consistency due to local cropping: the relocated car and inserted backpack lack the cast shadows and reflected illumination seen in \method{} and ground truth. For Remove, the baseline fails to eliminate the long shadow extending beyond its perspective crop, whereas \method{} removes the entire shadow.
}
\label{fig:pers_compare}
\end{figure*}

\subsection{Evaluation Metrics}

We report FAED \cite{wu2023panodiffusion}, PSNR, SSIM \cite{wang2004ssim}, LPIPS \cite{zhang2018lpips}, DINOv3 similarity \cite{simeoni2025dinov3}, and FID \cite{heusel2017fid}. FAED and PSNR are computed on the full ERP panorama: FAED measures panorama-level distributional quality with a panorama-specific autoencoder, and PSNR measures global reconstruction fidelity, so together they capture whether an editor preserves the scene outside the edit. The remaining metrics are computed on square perspective views centered on the edit region and rendered with the same adaptive field-of-view rule as the perspective protocol, because SSIM, LPIPS, DINOv3 features, and FID are designed for perspective image statistics and are easier to interpret in locally undistorted views. Higher is better for PSNR, SSIM, and DINOv3 similarity; lower is better for FAED, LPIPS, and FID.

\subsection{Quantitative Comparison}

\noindent\textbf{ERP protocol.}
Table~\ref{tab:main_results} reports results with all baselines applied directly on the ERP, separated into the UE5 and real test sets. On UE5, \method{} (bbox) obtains the best value on all six metrics for Translation (e.g., FID 95.5 vs.\ 108.9 for the strongest baseline) and on five of the six for Insert, where only the DINOv3 similarity of Insert-Anything (0.754 vs.\ 0.736) remains higher; even the lightweight point-guided variant surpasses every baseline on FAED, SSIM, LPIPS, and FID for Translation. Perspective baselines applied directly to ERP inputs also degrade panorama-level statistics: apart from Insert-Anything (0.344), their FAED is an order of magnitude worse than \method{} (1.2--8.8 vs.\ 0.31 for Translation on UE5), since editing distorted ERP content and regenerating entire panoramas both violate equirectangular image statistics. The advantage carries over to the real set: \method{} (bbox) obtains the best PSNR, SSIM, LPIPS, and FID for Translation and the best FAED, PSNR, SSIM, LPIPS, and FID for Insert, while Insert-Anything, built on a 12B FLUX-Fill backbone with inpainting-scale training data, keeps the best real FAED (0.229 vs.\ 0.356) and DINOv3 similarity for Translation and the best DINOv3 similarity for Insert. For Remove, \method{} attains the best value on all six metrics on UE5, with the mask variant leading FAED, SSIM, LPIPS, and FID and the bbox variant leading PSNR and DINOv3 similarity, and the best PSNR, SSIM, LPIPS, DINOv3 similarity, and FID on the real set; LaMa remains a strong specialized inpainter, retaining the best FAED on the real set (0.261 vs.\ 0.336), while SE360, the only public panorama-native competitor, trails on all metrics.

\noindent\textbf{Perspective protocol.}
Table~\ref{tab:main_results_pers} re-evaluates the baselines under the perspective protocol, which removes ERP distortion from their inputs and confines their edits to the projected region. As expected, this strictly favorable adaptation strengthens every baseline substantially, most visibly on panorama-level statistics on the real set: for Translation, the FAED of AnyDoor improves from 4.167 to 0.307 and that of OmniPaint from 2.340 to 0.245, because the panorama outside the re-projected edit region is preserved exactly. For fairness, \method{}'s outputs are composited identically: every pixel outside the perspective-view crop is replaced by the input panorama. Its FAED and PSNR therefore change relative to Table~\ref{tab:main_results} while its perspective-view metrics are unchanged. Even against these strengthened baselines, \method{} remains the strongest method on UE5 for all three tasks on FAED, PSNR, SSIM, LPIPS, and FID (and additionally on DINOv3 similarity for Remove), with only the DINOv3 similarity of Insert-Anything ahead for Translation and Insert. On the real set, \method{} leads all six metrics for Remove and every metric except DINOv3 similarity for Translation (e.g., FAED 0.185 vs.\ 0.214 for Insert-Anything), and keeps the best FAED, SSIM, and LPIPS for Insert, whereas Insert-Anything, the strongest FLUX-based editor, retains the best DINOv3 similarity for Translation and the best PSNR, DINOv3 similarity, and FID for Insert. We emphasize two structural advantages that the protocol cannot give the baselines: they require a known-size target bbox or mask and a staged remove-then-insert pipeline for Translation, whereas \method{} performs all three tasks in a single unified model and supports point-only guidance; and pinhole projection fundamentally cannot frame very large or near-field objects (about 5\% of samples exceed the $150^\circ$ limit and fall back to the ERP path), whereas \method{} operates on the full panorama natively.

\begin{table*}[t]
\centering
\scriptsize
\caption{Quantitative comparison under the \emph{ERP protocol}, where baselines operate directly on equirectangular panoramas, reported separately on the UE5 (210 samples per task) and real (50 samples per task) test sets. FAED and PSNR are computed on full ERP panoramas; SSIM, LPIPS, DINOv3 similarity, and FID are computed on perspective views centered on the edit region. \method{} (point) uses only a target point. All baselines receive the known-size target bbox or mask.}
\label{tab:main_results}
\resizebox{\textwidth}{!}{%
\begin{tabular}{ll cccccc cccccc}
\toprule
& & \multicolumn{6}{c}{UE5 test set (210)} & \multicolumn{6}{c}{Real test set (50)} \\
\cmidrule(lr){3-8}\cmidrule(lr){9-14}
Task & Method & FAED $\downarrow$ & PSNR $\uparrow$ & SSIM $\uparrow$ & LPIPS $\downarrow$ & DINOv3 $\uparrow$ & FID $\downarrow$ & FAED $\downarrow$ & PSNR $\uparrow$ & SSIM $\uparrow$ & LPIPS $\downarrow$ & DINOv3 $\uparrow$ & FID $\downarrow$ \\
\midrule
\multirow{7}{*}{Translation}
& DragAnything \cite{wu2024draganything} & 8.797 & 12.68 & 0.351 & 0.607 & 0.288 & 208.0 & 24.362 & 10.86 & 0.422 & 0.679 & 0.215 & 240.0 \\
& AnyDoor \cite{chen2024anydoor} & 1.157 & 22.18 & 0.571 & 0.352 & 0.687 & 116.0 & 4.167 & 22.00 & 0.673 & 0.345 & 0.611 & 167.0 \\
& Paint-by-Example \cite{yang2023paintbyexample} & 1.504 & 21.75 & 0.564 & 0.337 & 0.437 & 133.0 & 1.888 & 22.29 & 0.626 & 0.352 & 0.454 & 199.0 \\
& OmniPaint \cite{yu2025omnipaint} & 1.196 & 22.02 & 0.696 & 0.318 & 0.662 & 117.0 & 2.340 & 21.43 & 0.662 & 0.323 & 0.656 & 167.0 \\
& Insert-Anything \cite{song2025insertanything} & 0.344 & 24.23 & 0.809 & 0.297 & 0.739 & 108.9 & \textbf{0.229} & 26.23 & 0.802 & 0.279 & \textbf{0.718} & 151.0 \\
& \method{} (point) & 0.336 & 24.14 & 0.830 & 0.270 & 0.721 & 97.8 & 0.357 & 26.82 & 0.781 & 0.274 & 0.669 & 144.5 \\
& \method{} (bbox) & \textbf{0.314} & \textbf{25.08} & \textbf{0.836} & \textbf{0.255} & \textbf{0.747} & \textbf{95.5} & 0.356 & \textbf{26.85} & \textbf{0.823} & \textbf{0.249} & 0.704 & \textbf{131.0} \\
\midrule
\multirow{6}{*}{Insert}
& Paint-by-Example \cite{yang2023paintbyexample} & 1.060 & 22.41 & 0.671 & 0.306 & 0.480 & 133.0 & 1.243 & 23.95 & 0.643 & 0.347 & 0.504 & 183.0 \\
& AnyDoor \cite{chen2024anydoor} & 1.742 & 21.95 & 0.671 & 0.324 & 0.678 & 125.5 & 6.967 & 21.35 & 0.672 & 0.361 & 0.573 & 182.0 \\
& OmniPaint \cite{yu2025omnipaint} & 2.315 & 21.92 & 0.748 & 0.285 & 0.692 & 120.5 & 1.577 & 22.37 & 0.680 & 0.350 & 0.662 & 167.0 \\
& Insert-Anything \cite{song2025insertanything} & 0.416 & 23.95 & 0.800 & 0.305 & \textbf{0.754} & 109.1 & 0.382 & 26.99 & 0.755 & 0.315 & \textbf{0.722} & 162.5 \\
& \method{} (point) & 0.433 & 23.46 & 0.816 & 0.305 & 0.691 & 113.3 & 0.454 & 25.91 & 0.757 & 0.319 & 0.630 & 173.9 \\
& \method{} (bbox) & \textbf{0.369} & \textbf{24.52} & \textbf{0.837} & \textbf{0.259} & 0.736 & \textbf{101.5} & \textbf{0.354} & \textbf{27.04} & \textbf{0.788} & \textbf{0.265} & 0.652 & \textbf{161.0} \\
\midrule
\multirow{5}{*}{Remove}
& LaMa \cite{suvorov2022lama} & 0.517 & 26.64 & 0.921 & 0.147 & 0.745 & 91.1 & \textbf{0.261} & 29.88 & 0.873 & 0.176 & 0.723 & 122.0 \\
& OmniPaint \cite{yu2025omnipaint} & 0.654 & 25.10 & 0.887 & 0.180 & 0.714 & 91.7 & 0.294 & 28.25 & 0.835 & 0.191 & 0.724 & 123.4 \\
& SE360 \cite{zhong2025se360} & 3.423 & 22.81 & 0.836 & 0.261 & 0.543 & 126.0 & 7.782 & 18.13 & 0.730 & 0.336 & 0.501 & 198.5 \\
& \method{} (bbox) & 0.266 & \textbf{27.91} & 0.918 & 0.148 & \textbf{0.784} & 80.0 & 0.336 & 29.71 & 0.881 & 0.147 & \textbf{0.806} & 94.3 \\
& \method{} (mask) & \textbf{0.262} & 27.66 & \textbf{0.923} & \textbf{0.141} & 0.783 & \textbf{77.4} & 0.372 & \textbf{29.99} & \textbf{0.887} & \textbf{0.141} & 0.793 & \textbf{92.2} \\
\bottomrule
\end{tabular}}
\end{table*}

\begin{table*}[t]
\centering
\scriptsize
\caption{Quantitative comparison under the \emph{perspective protocol}, where each baseline edits a locally undistorted, adaptively framed square perspective view (FOV clipped to $[35^\circ,150^\circ]$) and only the edited region is re-projected into the panorama. This adaptation is strictly favorable to the baselines. \method{} operates natively on the ERP; for fairness, its output pixels outside the perspective-view crop are likewise replaced by the input panorama, so its FAED and PSNR differ from Table~\ref{tab:main_results} while the perspective-view metrics are unchanged. DragAnything is omitted because its video-based dragging does not transfer to this protocol.}
\label{tab:main_results_pers}
\resizebox{\textwidth}{!}{%
\begin{tabular}{ll cccccc cccccc}
\toprule
& & \multicolumn{6}{c}{UE5 test set (210)} & \multicolumn{6}{c}{Real test set (50)} \\
\cmidrule(lr){3-8}\cmidrule(lr){9-14}
Task & Method & FAED $\downarrow$ & PSNR $\uparrow$ & SSIM $\uparrow$ & LPIPS $\downarrow$ & DINOv3 $\uparrow$ & FID $\downarrow$ & FAED $\downarrow$ & PSNR $\uparrow$ & SSIM $\uparrow$ & LPIPS $\downarrow$ & DINOv3 $\uparrow$ & FID $\downarrow$ \\
\midrule
\multirow{6}{*}{Translation}
& AnyDoor \cite{chen2024anydoor} & 0.451 & 23.39 & 0.798 & 0.291 & 0.727 & 103.7 & 0.307 & 26.56 & 0.767 & 0.277 & 0.675 & 140.6 \\
& Paint-by-Example \cite{yang2023paintbyexample} & 0.559 & 23.30 & 0.817 & 0.302 & 0.513 & 131.6 & 0.459 & 25.42 & 0.768 & 0.303 & 0.561 & 176.4 \\
& Insert-Anything \cite{song2025insertanything} & 0.323 & 24.52 & 0.822 & 0.256 & \textbf{0.769} & 97.8 & 0.214 & 27.88 & 0.784 & 0.250 & \textbf{0.727} & 132.4 \\
& OmniPaint \cite{yu2025omnipaint} & 0.461 & 23.36 & 0.813 & 0.285 & 0.692 & 107.4 & 0.245 & 27.17 & 0.784 & 0.268 & 0.653 & 135.6 \\
& \method{} (point) & 0.265 & 24.25 & 0.830 & 0.270 & 0.721 & 97.8 & 0.294 & 26.04 & 0.781 & 0.274 & 0.669 & 144.5 \\
& \method{} (bbox) & \textbf{0.247} & \textbf{25.00} & \textbf{0.836} & \textbf{0.255} & 0.747 & \textbf{95.5} & \textbf{0.185} & \textbf{27.91} & \textbf{0.823} & \textbf{0.249} & 0.704 & \textbf{131.0} \\
\midrule
\multirow{6}{*}{Insert}
& Paint-by-Example \cite{yang2023paintbyexample} & 0.675 & 23.74 & 0.816 & 0.298 & 0.520 & 130.7 & 0.529 & 24.91 & 0.757 & 0.303 & 0.569 & 189.8 \\
& AnyDoor \cite{chen2024anydoor} & 0.495 & 23.31 & 0.791 & 0.298 & 0.729 & 109.7 & 0.343 & 26.50 & 0.760 & 0.283 & 0.700 & 140.1 \\
& Insert-Anything \cite{song2025insertanything} & 0.410 & 24.32 & 0.825 & 0.263 & \textbf{0.749} & 103.4 & 0.269 & \textbf{27.83} & 0.783 & 0.272 & \textbf{0.712} & \textbf{128.1} \\
& OmniPaint \cite{yu2025omnipaint} & 0.525 & 23.61 & 0.808 & 0.285 & 0.716 & 107.5 & 0.270 & 26.25 & 0.685 & 0.270 & 0.680 & 141.7 \\
& \method{} (point) & 0.403 & 23.69 & 0.816 & 0.305 & 0.691 & 113.3 & 0.414 & 26.01 & 0.757 & 0.319 & 0.630 & 173.9 \\
& \method{} (bbox) & \textbf{0.375} & \textbf{24.71} & \textbf{0.837} & \textbf{0.259} & 0.736 & \textbf{101.5} & \textbf{0.265} & 27.79 & \textbf{0.788} & \textbf{0.265} & 0.652 & 161.0 \\
\midrule
\multirow{4}{*}{Remove}
& LaMa \cite{suvorov2022lama} & 0.582 & 26.57 & 0.915 & 0.174 & 0.695 & 96.2 & 0.296 & 30.12 & 0.883 & 0.200 & 0.749 & 119.9 \\
& OmniPaint \cite{yu2025omnipaint} & 0.555 & 25.23 & 0.875 & 0.219 & 0.625 & 105.5 & 0.342 & 29.03 & 0.849 & 0.203 & 0.680 & 120.1 \\
& \method{} (bbox) & 0.296 & \textbf{28.22} & 0.918 & 0.148 & \textbf{0.784} & 80.0 & 0.216 & 30.66 & 0.881 & 0.147 & \textbf{0.806} & 94.3 \\
& \method{} (mask) & \textbf{0.288} & 28.06 & \textbf{0.923} & \textbf{0.141} & 0.783 & \textbf{77.4} & \textbf{0.204} & \textbf{31.18} & \textbf{0.887} & \textbf{0.141} & 0.793 & \textbf{92.2} \\
\bottomrule
\end{tabular}}
\end{table*}

\begin{table*}[t]
\centering
\scriptsize
\caption{Ablation study on the auxiliary depth condition and the spatial-instruction granularity (point/bbox/mask), reported separately on the UE5 and real test sets. FAED and PSNR are computed on full ERP panoramas; the remaining metrics are computed on perspective views centered on the edit region. The w/o-depth variant uses point guidance for Translation and Insert and bbox guidance for Remove.}
\label{tab:ablation_results}
\resizebox{\textwidth}{!}{%
\begin{tabular}{ll cccccc cccccc}
\toprule
& & \multicolumn{6}{c}{UE5 test set (210)} & \multicolumn{6}{c}{Real test set (50)} \\
\cmidrule(lr){3-8}\cmidrule(lr){9-14}
Task & Method & FAED $\downarrow$ & PSNR $\uparrow$ & SSIM $\uparrow$ & LPIPS $\downarrow$ & DINOv3 $\uparrow$ & FID $\downarrow$ & FAED $\downarrow$ & PSNR $\uparrow$ & SSIM $\uparrow$ & LPIPS $\downarrow$ & DINOv3 $\uparrow$ & FID $\downarrow$ \\
\midrule
\multirow{4}{*}{Translation}
& \method{} w/o depth & 0.351 & 24.42 & 0.813 & 0.275 & 0.708 & 100.0 & 0.374 & 26.53 & 0.752 & 0.297 & 0.655 & 148.5 \\
& \method{} (point) & 0.336 & 24.14 & 0.830 & 0.270 & 0.721 & 97.8 & 0.357 & 26.82 & 0.781 & 0.274 & 0.669 & 144.5 \\
& \method{} (bbox) & 0.314 & 25.08 & 0.836 & 0.255 & 0.747 & 95.5 & 0.356 & 26.85 & 0.823 & 0.249 & 0.704 & 131.0 \\
& \method{} (mask) & \textbf{0.250} & \textbf{25.43} & \textbf{0.868} & \textbf{0.197} & \textbf{0.780} & \textbf{82.9} & \textbf{0.186} & \textbf{29.35} & \textbf{0.841} & \textbf{0.181} & \textbf{0.717} & \textbf{121.6} \\
\midrule
\multirow{4}{*}{Insert}
& \method{} w/o depth & 0.438 & 23.42 & 0.821 & 0.355 & 0.671 & 112.0 & 0.493 & 25.59 & 0.734 & 0.335 & 0.625 & 174.1 \\
& \method{} (point) & 0.433 & 23.46 & 0.816 & 0.305 & 0.691 & 113.3 & 0.454 & 25.91 & 0.757 & 0.319 & 0.630 & 173.9 \\
& \method{} (bbox) & 0.369 & 24.52 & 0.837 & 0.259 & 0.736 & 101.5 & 0.354 & 27.04 & 0.788 & 0.265 & 0.652 & 161.0 \\
& \method{} (mask) & \textbf{0.332} & \textbf{24.76} & \textbf{0.868} & \textbf{0.199} & \textbf{0.775} & \textbf{88.6} & \textbf{0.226} & \textbf{29.00} & \textbf{0.829} & \textbf{0.193} & \textbf{0.732} & \textbf{129.2} \\
\midrule
\multirow{3}{*}{Remove}
& \method{} w/o depth & 0.284 & 27.64 & 0.846 & 0.153 & 0.744 & 80.0 & \textbf{0.326} & 29.02 & 0.868 & 0.164 & 0.793 & 95.4 \\
& \method{} (bbox) & 0.266 & \textbf{27.91} & 0.918 & 0.148 & \textbf{0.784} & 80.0 & 0.336 & 29.71 & 0.881 & 0.147 & \textbf{0.806} & 94.3 \\
& \method{} (mask) & \textbf{0.262} & 27.66 & \textbf{0.923} & \textbf{0.141} & 0.783 & \textbf{77.4} & 0.372 & \textbf{29.99} & \textbf{0.887} & \textbf{0.141} & 0.793 & \textbf{92.2} \\
\bottomrule
\end{tabular}}
\end{table*}

\subsection{Qualitative Comparison}
\label{sec:qualitative}

Qualitatively, perspective-image baselines often struggle with panorama-specific artifacts: for Translation, they may distort the moved object or leave residual source content; for Insert, they may copy viewpoint-specific reference appearance; and for Remove, specialized inpainters can produce clean textures but are not designed for the unified manipulation setting.
For qualitative inspection, we use both full ERP outputs and perspective crops. Full ERP panoramas are necessary for checking global scene preservation and left-right boundary continuity, while perspective crops make it easier to inspect object geometry, contact with support surfaces, local lighting, and residual artifacts. Figures~\ref{fig:move_compare}--\ref{fig:remove_compare} compare \method{} with representative baselines for the three tasks under the ERP protocol.

Figure~\ref{fig:pers_compare} examines the perspective protocol qualitatively. Although editing in a locally undistorted crop removes ERP distortion from the baseline's input, two structural limits remain: the mask-conditioned baselines can synthesize content only inside the provided target box, and the scene context they observe ends at the crop boundary. For Translation and Insert, the baselines place the object at the requested location, but the relocated car and the inserted backpack lack the cast shadows and reflected light that both \method{} and the ground truth exhibit, since a correct shadow would have to be painted outside the target box. For Remove, even though the entire edited view is re-projected, the crop bounds what the baseline can repair: OmniPaint removes the street light and the shadow segment inside its view, but the long cast shadow sweeping across the plaza extends far beyond any pinhole view of the lamp and survives in the output. \method{}, editing the full panorama natively, removes the object together with its entire shadow and keeps illumination globally consistent.

\subsection{Ablation Study}
\label{sec:ablation}

Table~\ref{tab:ablation_results} reports ablation results, separated into the UE5 and real test sets. Two factors are ablated: the auxiliary depth condition and the granularity of the spatial instruction (point, bbox, or mask). The depth ablation compares two separately trained models, whereas the three guidance granularities share the same trained weights and differ only in the instruction supplied at inference.

\noindent\textbf{Auxiliary depth.}
Removing the DA$^2$ depth condition degrades every Translation and Insert metric on the real test set (e.g., real Translation DINOv3 similarity drops from 0.669 to 0.655) and most metrics on UE5, where only UE5 Translation PSNR and UE5 Insert SSIM and FID marginally favor the w/o-depth variant, confirming that predicted depth is a useful auxiliary geometric cue rather than a dominant supervision signal. For Remove, depth is also beneficial: the full model matches or improves every UE5 metric, most visibly SSIM (0.918 vs.\ 0.846), and improves every real metric except FAED, where the w/o-depth variant is marginally better (0.326 vs.\ 0.336). Although DA$^2$ estimates depth from the unedited input, so the depth condition inside the source region still encodes the removed object, the surrounding depth context evidently helps the model complete the background consistently. Figure~S4 in the supplementary material shows this benefit qualitatively under the default point guidance, where the model must infer the projected size of the relocated object on its own: without depth, the object is rendered at an implausibly large scale for the target location, whereas the depth-conditioned model produces a size consistent with the local scene depth. Overall, auxiliary depth benefits all three tasks, with its largest single effect on the structural fidelity (SSIM) of Remove.

\noindent\textbf{Instruction granularity.}\looseness=-1
For Translation and Insert, guidance granularity improves results monotonically on every metric and both test sets: point $\rightarrow$ bbox $\rightarrow$ mask. The bbox variant adds an explicit target-size prior over the point variant, and the mask variant further specifies the target silhouette, yielding the best values on every metric (e.g., real Insert FAED improves from 0.454 with a point to 0.354 with a bbox and 0.226 with a mask). For Remove, the bbox and mask variants are close: mask is consistently better on SSIM, LPIPS, and FID, bbox retains slightly higher DINOv3 similarity on both sets, and FAED and PSNR split between the two. Notably, the gap between guidance levels is larger on the real set than on UE5, suggesting that explicit spatial priors compensate for domain shift. Point guidance remains the default lightweight interface as it requires the least user input, and even this weakest variant is competitive with the strongest baselines in Tables~\ref{tab:main_results} and \ref{tab:main_results_pers}.

\raggedbottom

\begin{figure}[t]
\centering
\includegraphics[width=\columnwidth]{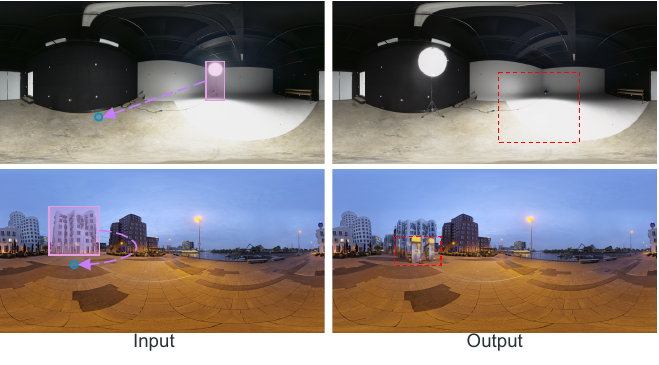}
\caption{Failure cases of \method{}. Top: translating a large-area light source. The emitter is relocated to the target point, but the wide illumination it casts is not re-synthesized: the brightly lit region anchored to the original position survives (dashed boxes), decoupling the moved source from the scene lighting; compact emitters do not trigger this failure. Bottom: translating a building-scale structure. The building is relocated, but additional structures absent from the input are hallucinated near the target (dashed box).}
\label{fig:failure}
\end{figure}

\subsection{Discussion and Limitations}
\label{sec:limitations}

\method{} currently uses three fixed task prompts. This preserves the pretrained text-conditioning interface but does not yet provide open-ended language control such as changing object attributes or specifying complex relational constraints. The default point-guided interface is lightweight, but it can be ambiguous when multiple support surfaces are near the target point or when the desired object scale is unusual. Bbox guidance provides stronger size control, but requires more user input. Predicted depth can also fail in reflective, transparent, textureless, or strongly distorted regions, and the model may struggle with complex cast shadows, reflections, or object-object interactions. Finally, although lighting randomization and real evaluation reduce the synthetic-to-real gap, UE5 assets and rendered lighting may still differ from real panoramas.

Figure~\ref{fig:failure} shows two characteristic failure modes that bound the scale of edits the current model can handle. The first concerns light sources with a large emitting area. Compact emitters such as a lamp can be translated together with their local illumination, but for a large-area source a correct edit amounts to relighting most of the panorama: \method{} relocates the emitter itself, while the wide pool of light it casts on the surrounding surfaces stays anchored to the original position, so the moved source and the scene illumination become decoupled. The second concerns scene-level structures. When asked to translate a building, the model relocates the building itself, but additional structures that do not exist in the input are hallucinated near the target region. Both edits lie far outside the training distribution of our UE5 corpus, whose motion paths displace object-scale assets and never require relighting an entire scene or moving architecture. Extending the corpus with such cases is a natural direction for future work.

\section{Conclusion}
\label{sec:conclusion}

We presented \method{}, a translation-centered framework for controllable object manipulation in 360$^\circ$ ERP panoramas, with Remove and reference-guided Insert supported as auxiliary tasks by the same model. Its two core ingredients are an interaction design and a paired data construction. The former is a three-channel ERP-aligned instruction map that unifies point, bbox, and mask guidance, so that a single click suffices to relocate an object; the latter is a UE5 pipeline that renders surface-aware object trajectories at scale, plus a dual-domain benchmark whose real captured tuples provide ground truth for all three tasks. The model itself is a purpose-designed multi-condition adaptation of a pretrained rectified-flow transformer (LoRA fine-tuning, ERP-aligned condition latents, and auxiliary depth conditioning), with the backbone unchanged. Across both test domains and two evaluation protocols, \method{} achieves strong reconstruction fidelity, semantic consistency, and distributional quality while supporting both lightweight point guidance and explicit bbox- or mask-based control.




\balance

\bibliographystyle{abbrv-doi-hyperref}
\bibliography{sample-bibliography}


\makeatletter
\setlength{\@dblfpsep}{14pt plus 2pt}
\makeatother

\setcounter{section}{0}
\setcounter{table}{0}
\setcounter{figure}{0}
\renewcommand{\thesection}{S\arabic{section}}
\renewcommand{\thetable}{S\arabic{table}}
\renewcommand{\thefigure}{S\arabic{figure}}
\renewcommand{\theHsection}{supp.\arabic{section}}
\renewcommand{\theHtable}{supp.\arabic{table}}
\renewcommand{\theHfigure}{supp.\arabic{figure}}

\twocolumn[{%
  \centering
  {\Large\bfseries Supplementary Material for\\[2pt]
   \method{}: Controllable Object Manipulation in 360$^\circ$ Panoramic Images\par}
  \vspace{1.5\baselineskip}
  {\small
   \captionof{table}{Adaptation protocols used to evaluate perspective object-editing baselines on ERP panoramas. All baselines receive the required known-size target bbox or mask, whereas \method{} is evaluated by default with only a target point.}
   \label{tab:baseline_protocols}
   \begin{tabular}{llp{4.6cm}p{6.9cm}}
   \toprule
   Method & Task(s) & Native interface & Protocol on ERP panoramas \\
   \midrule
   DragAnything \cite{wu2024draganything} & Translation & Drag handles with video trajectory & Source handle at object center, target handle at desired location; linear trajectory; last generated video frame used as the edited panorama. \\
   AnyDoor \cite{chen2024anydoor} & Translation, Insert & Reference image + target mask/bbox & Insert: reference insertion at the known-size target bbox. Translation: remove-then-insert, with OmniPaint removal and the cropped source object as reference. \\
   Paint-by-Example \cite{yang2023paintbyexample} & Translation, Insert & Reference image + target mask & Same insertion and remove-then-insert protocol as AnyDoor. \\
   OmniPaint \cite{yu2025omnipaint} & Translation, Remove, Insert & Object mask (removal); reference + target region (insertion) & Native removal and insertion; Translation via removal followed by insertion of the cropped source object at the target region. \\
   Insert-Anything \cite{song2025insertanything} & Translation, Insert & Reference image + mask (FLUX-Fill diptych inpainting) & Insert: reference insertion at the known-size target bbox. Translation: remove-then-insert, with OmniPaint removal and the source object as reference. \\
   LaMa \cite{suvorov2022lama} & Remove & Inpainting mask & Source-object mask inpainting on the ERP panorama. \\
   SE360 \cite{zhong2025se360} & Remove & Text prompt + ERP instruction mask & Native ERP removal with a united task prompt; spatial control from the bbox instruction mask (no object names are available). \\
   \bottomrule
   \end{tabular}\par}
  \vspace{1.5\baselineskip}
}]

\section{Baseline Adaptation Protocols}
\label{sec:supp_baseline_protocols}

Since all compared editors are designed for perspective images, we adapt each baseline to its native interface, as described in Section~\ref{sec:baselines} of the main paper. Table~\ref{tab:baseline_protocols} summarizes the per-method protocols.

\section{Additional Qualitative Results}
\label{sec:supp_qualitative}

\begin{figure*}[t]
\centering
\includegraphics[width=\textwidth]{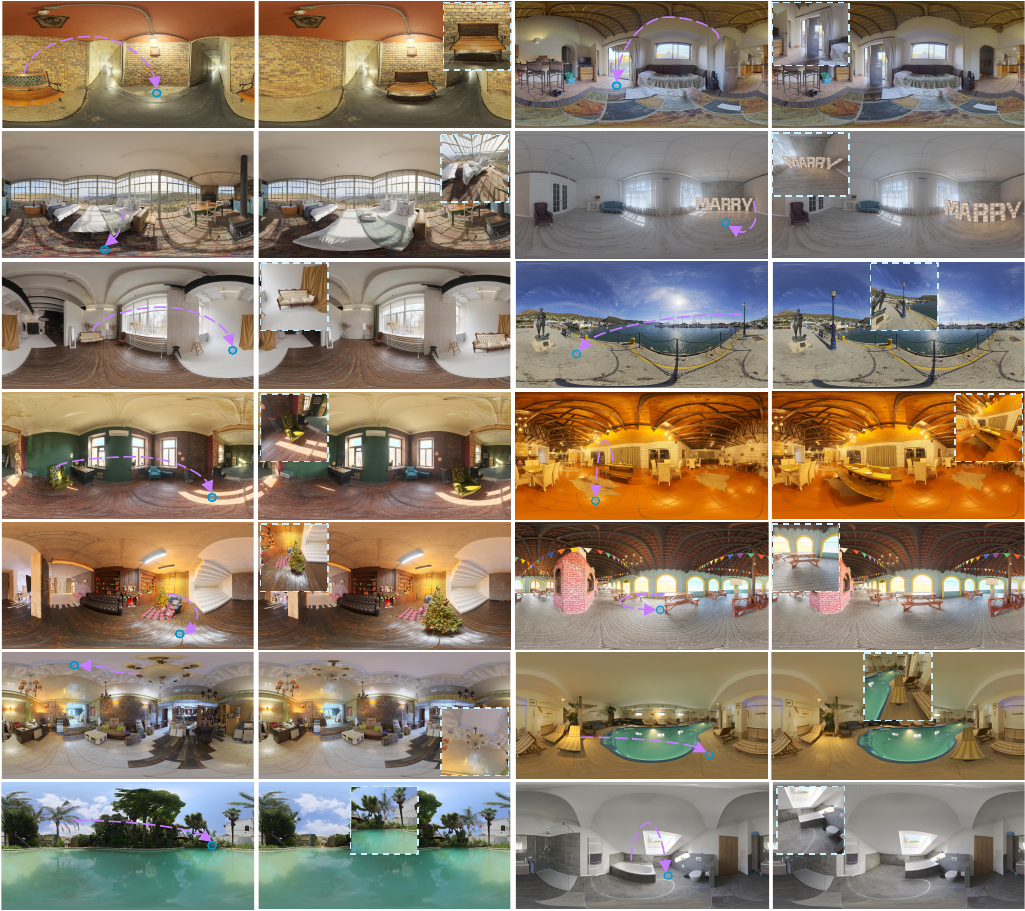}
\caption{Additional Translation results. Each pair shows the input panorama with the source region, the motion arrow, and the Gaussian target point overlaid, and the edited panorama with a magnified view of the target region (dashed box).}
\label{fig:supp_translation}
\end{figure*}

\begin{figure*}[t]
\centering
\includegraphics[width=0.85\textwidth]{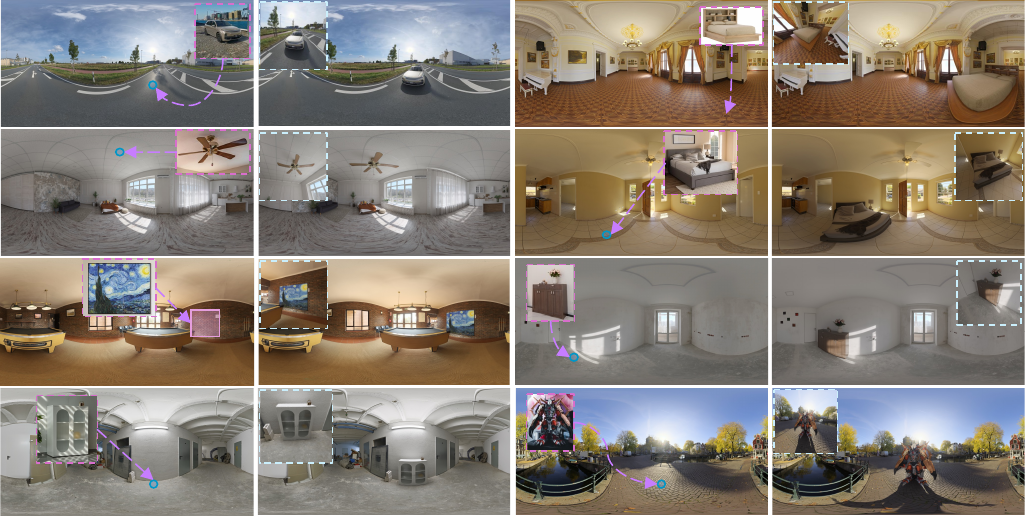}
\caption{Additional Insert results. Each pair shows the input panorama with the reference object (inset) and the target point overlaid, and the edited panorama with a magnified view of the inserted object (dashed box).}
\label{fig:supp_insert}
\end{figure*}

\begin{figure*}[t]
\centering
\includegraphics[width=0.85\textwidth]{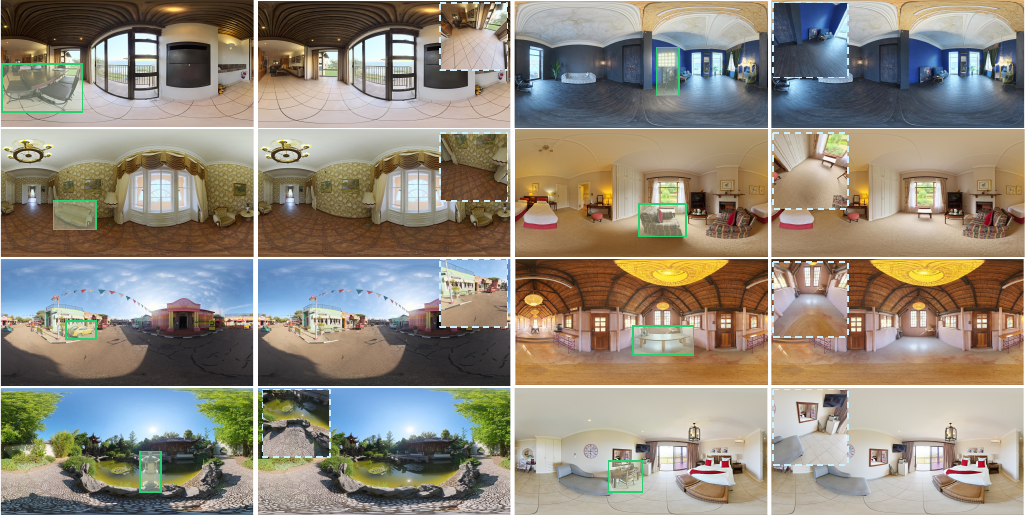}
\caption{Additional Remove results. Each pair shows the input panorama with the source object marked, and the edited panorama with a magnified view of the completed background (dashed box).}
\label{fig:supp_remove}
\end{figure*}

Figures~\ref{fig:supp_translation}--\ref{fig:supp_remove} show additional 360$^\circ$ object manipulation results produced by \method{} for the Translation, Insert, and Remove tasks, complementing the qualitative comparisons in the main paper. Each pair shows the input panorama with the instruction overlaid and the edited panorama with a magnified view of the edited region (dashed box).

\begin{figure*}[t]
\centering
\includegraphics[width=0.9\textwidth]{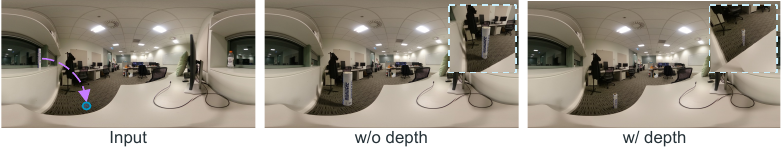}
\caption{Qualitative ablation of the auxiliary depth condition under the default point guidance. Asked to relocate the object to the target point (arrow), the w/o-depth model renders it at an implausibly large scale, whereas the depth-conditioned model infers a projected size consistent with the scene depth at the target location (insets).}
\label{fig:supp_ablation_qual}
\end{figure*}

Figure~\ref{fig:supp_ablation_qual} further provides a qualitative view of the auxiliary-depth ablation of the main paper (Section~\ref{sec:ablation}) under the default point guidance, where the model must infer the projected size of the relocated object on its own: without depth, the object is rendered at an implausibly large scale for the target location, whereas the depth-conditioned model produces a size consistent with the local scene depth.

\end{document}